%% file: clue.tex
\documentclass[11pt]{article}
\usepackage{acl2010}
\usepackage[numbers,square,sort&compress,longnamesfirst]{natbib}
\usepackage{times}
\usepackage{latexsym}
\usepackage{xspace}
\usepackage{graphicx}
\usepackage[hyperfootnotes=false]{hyperref}

\newif{\ifshort}
\shortfalse

\newif{\ifepoint}
\epointtrue

\hypersetup{ 
colorlinks,
citecolor=black,
filecolor=black,
linkcolor=black,
urlcolor=black} 
\usepackage[dvipsnames]{color}
\usepackage{multicol}
\usepackage[overload]{textcase}
\usepackage{amssymb}
\usepackage{centernot}
\usepackage{enumitem}
\usepackage{setspace}

\usepackage{tikz}
\usetikzlibrary{arrows}
\usetikzlibrary{decorations.pathreplacing}

\input{nsub-macros}

\input{acl-short-macros}

\title{Don't `have a clue'?  
\\ Unsupervised co-learning of downward-entailing operators
\ifshort{}\else{\\ \vspace*{-.09in}{\small \textcolor{red}{%
\fbox{pp. 1-6 are identical to the ACL 2010 published version; pp. 7-8 are the ``externally-available appendices''. }}}\\
\vspace*{-.1in}}\fi
}
\author{Cristian Danescu-Niculescu-Mizil and Lillian Lee \\ Department of Computer Science,  Cornell University \\
{cristian@cs.cornell.edu, llee@cs.cornell.edu}}

\begin{document}

\maketitle
\input{acl-short-intro}

\input{acl-short-method}

\input{acl-short-results}

\input{acl-short-multilingual}

\input{acl-short-conclusions}

\input{esub-ack}

\bibliographystyle{plainnat}

\ifshort
\end{document}

%% file: nsub-macros.tex
\newcommand{\ex}[1]{`{\it #1}'}

\newcommand{\dm}{DE\xspace}

\newcommand{\monotonic}{entailing}
\newcommand{\dmlong}{downward \monotonic\xspace}
\newcommand{\dmlonghyph}{downward-\monotonic\xspace}
\newcommand{\Dmlonghyph}{Downward-\monotonic\xspace}
\newcommand{\umlong}{upward \monotonic\xspace}
\newcommand{\umlonghyph}{upward-\monotonic\xspace}
\newcommand{\oplong}{operator\xspace}
\newcommand{\opslong}{operators\xspace}
\newcommand{\po}{\dm \oplong}%
\newcommand{\polong}{\dmlonghyph \oplong}%
\newcommand{\pos}{{\po}s\xspace}
\newcommand{\Pos}{{\po}s\xspace}

\newcommand{\polongs}{{\polong}s\xspace}

%% file: acl-short-macros.tex
\newcommand{\mynumberfig}[1]{\setcounter{figure}{#1}\addtocounter{figure}{-1}}
\newcommand{\numfigbarsnoiter}{3} %
\newcommand{\numfigenglish}{5}
\newcommand{\mynumbersubsec}[1]{\setcounter{subsection}{#1}\addtocounter{subsection}{-1}}
\newcommand{\numsecmarginal}{2}

\newcommand{\shrink}[1]{\ifepoint{}\else#1\fi}

\newcommand{\unshrink}[1]{#1} 

\newcommand{\dld}{DLD09\xspace} %

\renewcommand{\ex}[1]{`{#1}'}
\newcommand{\showop}[1]{\fbox{#1}}
\newcommand{\showupwardop}[1]{\underline{#1}}

\newcommand{\tueb}{CoDII-NPI.ro\xspace}

\newcommand{\scorepos}{f}
\newcommand{\scorepnpis}{\scorepos_r}

\newcommand\pdfsite{http://www.cs.cornell.edu/~cristian/acl2010/paper/}
\newcommand{\elsewhere}{\ifshort\hypersetup{urlcolor=blue}\else\fi
in the %
\href{\pdfsite}{externally-available appendices}\hypersetup{urlcolor=black}\xspace}

\urldef\mysite\url{http://www.cs.cornell.edu/~cristian/acl2010/}

\newcommand{\elsewhereurl}{\ifshort\hypersetup{urlcolor=blue}\else\fi
\href{\pdfsite}{{www.cs.cornell.edu/\string~cristian/acl2010/}}\hypersetup{urlcolor=black}}

\newcommand{\dataurl}{\hypersetup{urlcolor=blue}\href{http://www.cs.cornell.edu/\%7Ecristian/acl2010/}{{http://www.cs.cornell.edu/\string~cristian/acl2010/}}\hypersetup{urlcolor=black}}

%% file: acl-short-intro.tex
\begin{abstract}
Researchers in textual entailment have begun to consider inferences
involving {\em \polongs}, an interesting and important class of
lexical items that change the way inferences are made. Recent work
proposed a method for learning English \polongs that requires access
to a high-quality collection of {\em negative polarity items} (NPIs).
However, English is one of the very few languages for which such a
list exists.  We propose the first approach that can be applied to the
many languages for which there is no pre-existing high-precision
database of NPIs.  As a case study, we apply our method to Romanian
and show that our method yields good results.  Also, we perform a
cross-linguistic analysis that suggests interesting connections to
some findings in linguistic typology.

\end{abstract}

\section{Introduction}

{\small{\it
\begin{tabular}{ll}
Cristi: & ``Nicio" ... is that adjective you've  mentioned. \\
Anca: & A negative pronominal adjective. \\
Cristi: &You mean there are people who analyze  that \\ & kind of thing? \\
Anca: &  The Romanian Academy. \\
Cristi: &  They're crazy. \\
\end{tabular}}

\hfill ---From the movie {\em Police, adjective}
}
\medskip

\Dmlonghyph \shrink{(\dm)} \opslong\ {are an interesting and varied class of
lexical items that   change the 
default way of dealing with  certain types of inferences.
 They  thus} play an important role 
in 
understanding natural language 
\citep[etc.]{vanBenthem:86a,SanchezValencia:91a,Dowty:94a,vanderWouden:97a}.

We explain what
{\dmlong} 
means by first  demonstrating the ``default" behavior, which is {\em \umlong}.  The word \ex{observed} is an example \umlonghyph\oplong:
the statement
\newcommand{\scrunchhere}{\vspace*{-.09in}}
\vspace*{-.2in}
\begin{enumerate}[label=(\roman*)]
\item \ex{Witnesses \showupwardop{observed} opium use.} 
\end{enumerate} 
\scrunchhere
implies 
\scrunchhere
\begin{enumerate}[label=(\roman*),start=2]
\item  \ex{Witnesses \showupwardop{observed} narcotic use.}
 \end{enumerate}
 but not vice versa (we write i $\Rightarrow (\centernot{\Leftarrow})$ ii).  
That is, the truth value is preserved if we replace the argument of an \umlonghyph \oplong by
 a superset (a more general version);  in our case,
the set  \ex{opium use} was replaced by the superset \ex{narcotic use}.

{\em \Dmlonghyph (\dm)} 
(also known  as  {\em downward monotonic} 
or {\em monotone decreasing})  \opslong
violate this default inference rule\unshrink{:
with \pos, reasoning instead goes from ``sets to subsets"}.
An example is the word \ex{bans}:
\begin{quote}\ex{The law \showop{bans} opium use} \\ $\centernot\Rightarrow (\Leftarrow)$\\ \ex{The law \showop{bans} narcotic use}.\end{quote}
Although \dm behavior represents an exception to the 
default, 
\pos are as a class rather common.
They are also quite diverse in sense and even part of speech. 
Some are simple negations,  such as \ex{not},
but some other English \pos  are
\ex{without}, \ex{reluctant to},  \ex{to doubt}, and \ex{to allow}.\footnote{Some examples showing different constructions for analyzing these \opslong:  \ex{The defendant does {not} own a blue car}  $\centernot\Rightarrow (\Leftarrow)$ \ex{The defendant does not own a car};   \ex{They are reluctant to tango}  $\centernot\Rightarrow (\Leftarrow)$ \ex{They are reluctant to dance};  \ex{Police doubt Smith threatened Jones}   $\centernot\Rightarrow (\Leftarrow)$ \ex{Police doubt Smith threatened Jones or Brown};  \ex{You are allowed to use Mastercard}   $\centernot\Rightarrow (\Leftarrow)$ \ex{You are allowed to use any credit card}. }
This variety makes
them hard to extract
automatically.

Because \pos
 violate the 
default
 ``sets to supersets'' inference,
identifying them can potentially improve performance in many
NLP tasks\unshrink{.
Perhaps the most obvious such tasks are those involving textual
entailment}, such as 
question answering, information extraction, summarization, and the evaluation of machine translation \citep{Dagan+Glickman+Magnini:06a}.
Researchers are in fact beginning to build textual-entailment systems that
can handle inferences involving \polongs other than simple negations,
although  these systems almost all rely on small handcrafted lists of \pos
\citep{Nairn+Condoravdi+Kartunnen:06a,MacCartney+Manning:08a,Christodoulopoulos:08a,Bar-Haim+al:08a,Breck:09a}.\footnote{The exception \citep{Breck:09a} employs the list automatically derived by \citet{Danescu-Niculescu-Mizil+Lee+Ducott:09a}, described later.}
Other
application areas are natural-language generation and human-computer interaction, 
since downward-entailing inferences
induce greater cognitive load than inferences in the opposite
direction \citep{Geurts+vanDerSlik:05a}.

\unshrink{Most NLP systems for the applications mentioned above have only been deployed for a small subset of languages.  
A 
key factor is the lack of relevant resources  for other languages. While one approach would be to separately develop a method to acquire such resources for each
language
individually, we instead  aim to ameliorate the resource-scarcity problem 
in the case of \pos
wholesale:
we propose
a single 
unsupervised
method that can
 extract  \pos
in
any language
for which 
raw text
corpora exist. 
}

{\paragraph{Overview of our work}

{
\renewcommand{\ex}[1]{{\it #1}}
\newcommand{\poex}[1]{\showop{#1}}%
\newcommand{\npiex}[1]{{\it #1}}
\newcommand{\bad}{$\times$\xspace}
\begin{figure*}[t]
\begin{center}
{\small}
\begin{tabular}{r|ll}
   & \multicolumn{2}{c}{{\bf NPIs}} \\ 
 {\bf \pos}      & \multicolumn{1}{c}{$\npiex{any}^{5}$} &
\multicolumn{1}{c}{\npiex{\npiex{have a clue}}, idiomatic sense} \\  \cline{2-3}%
\poex{not} or \poex{n't}  & $\checkmark$ We do\poex{n't} have \npiex{any} apples & $\checkmark$ We  do\poex{n't} \npiex{have a clue} \\ 
\poex{doubt}			& \checkmark I \poex{doubt} they have \npiex{any}
apples   & $\checkmark$ I \poex{doubt} they \npiex{have a clue} \\     %
no \po        &  \bad They have \npiex{any} apples & \bad They \npiex{have a clue} \\
\end{tabular}
\end{center}
\caption{\label{fig:ladexamples} Examples consistent with Ladusaw's hypothesis that NPIs can only occur within the scope of \pos.  A $\checkmark$ denotes an  acceptable sentence; a $\times$ denotes an unacceptable sentence. %
}
\end{figure*}
}

Our approach takes the English-centric work  of \citet{Danescu-Niculescu-Mizil+Lee+Ducott:09a} --- \dld for short ---  as a starting point, as 
they present
 the first and, until now, only algorithm for automatically extracting \pos from data.  However, \unshrink{our work departs significantly from \dld in the following  key respect.

}\dld critically depends on 
access to
a high-quality, carefully curated collection of 
{\em
  negative polarity items (NPIs)} --- lexical items such as
\ex{any},
\ex{ever},
or 
the idiom \ex{have a clue} 
that tend to occur only in negative
environments
(see 
\S \ref{sec:dld} for more 
details).
\dld 
use
NPIs
as signals of the occurrence of 
\polongs.
However, 
 almost every language other than English lacks a high-quality accessible NPI list.

{
To circumvent this \shrink{widespread }problem\shrink{ of lack of NPI lists}, we introduce a
knowledge-lean  {\em co-learning} approach\shrink{ to
the unsupervised discovery of \polongs}. }
Our algorithm is initialized with a very small seed set of NPIs
 (which we describe how to 
generate),
  and then iterates between (a) discovering a set of \pos using a collection of {\em pseudo-NPIs}
  --- a concept we introduce ---
  and
(b) using the newly-acquired \pos to detect new pseudo-NPIs.

{\vspace*{-.05in}}

\paragraph{Why this isn't obvious} 
Although the algorithmic idea sketched above 
seems quite simple, it is important to note that {prior experiments in that direction have not proved fruitful.}  
{Preliminary} work on learning
(German) NPIs using a small list of
simple known  \pos did not yield strong results 
\citep{Lichte+Soehn:07a}. 
\citet{Hoeksema:97a}
discusses
why NPIs might be hard to learn from data.
\footnote{In fact, 
humans can have trouble agreeing on NPI-hood;
for 
instance,
\citet{Lichte+Soehn:07a} mention doubts about over half of
\citet{Kuerschner:83a}'s 344 manually collected
German NPIs.}
We circumvent this problem because we are not interested in learning NPIs per se; rather, for our purposes, pseudo-NPIs suffice.
{Also}, our preliminary work determined that  one of the most famous co-learning algorithms,   {\em hubs and authorities} or {\em HITS} \citep{Kleinberg:98a}, is poorly suited to our problem.\footnote{
{We} explored three different edge-weighting schemes based on co-occurrence frequencies and seed-set membership, but the results were extremely poor;  HITS invariably retrieved very frequent words.
}

\paragraph{Contributions}

To begin with, 
we apply our algorithm to produce the first large list of \pos for a language other than English.
In our case study 
on
Romanian
(\S \ref{sec:results}), 
 we achieve quite high precisions at $k$
(for example, iteration achieves a precision at 30 of 87\%).

Auxiliary experiments explore the effects of using a large but noisy NPI list, should one be available for the language in question. 
 Intriguingly, we find that co-learning new pseudo-NPIs provides better results.

Finally (\S\ref{sec:multilingual}), we engage in some cross-linguistic analysis 
based on the results of applying  our algorithm to English. 
We find
that
 there are some suggestive connections with 
findings in linguistic typology.

\paragraph{Appendix available}  A more complete account of our work and its implications can be found in  a version of this paper containing appendices, available at 
\elsewhereurl.

%% file: acl-short-method.tex
\section{%
\dld: successes and challenges}\label{sec:dld}

In this section, we briefly summarize those aspects of the \dld
method that are important to understanding how our 
new  co-learning method works.

\paragraph{\Pos and NPIs}
Acquiring \pos is challenging because of the 
complete
lack of annotated data.
\dld's insight was to make 
use of {\em negative polarity items
(NPIs)},
which are words or phrases that tend to occur only in negative
contexts.  The reason they 
did
so is that  Ladusaw's  hypothesis \citep{Fauconnier:75a,Ladusaw:80a} asserts that {\em NPIs only occur within the scope of \pos}. 
 Figure \ref{fig:ladexamples} depicts examples involving the 
English NPIs \ex{any}\footnote{The 
{\em free-choice} sense of \ex{any}, as in \ex{I
  can skim any paper in five minutes}, is a known exception.}  and \ex{have a clue} 
(in the idiomatic sense)  
  that 
illustrate this relationship.  
Some other English NPIs are \ex{ever}, \ex{yet} and \ex{give a damn}.

Thus, NPIs can be treated as clues that a \po
 might be present (although \pos
 may also occur without NPIs).

\paragraph{\dld algorithm}
Potential \pos are collected by extracting those words that
 appear in an NPI's 
context
 at least once.\footnote{\dld policies: (a) 
 ``NPI context'' was defined as the part of the sentence to the left
of the NPI
 up to the first comma, 
 semi-colon or beginning of sentence;  (b) to encourage the discovery of new \pos, those sentences containing one of a list of 10 well-known \pos were discarded.  
For Romanian, we 
treated
only negations (\ex{nu} and \ex{n-}) and questions as well-known 
environments.
 } Then, the potential 
 \pos $x$ are ranked by 

{\small \vspace*{-.2in}
$$ \scorepos(x) :=  \frac
{\mbox{fraction of NPI contexts that contain $x$} }
{\mbox{relative frequency of $x$ in the corpus}},
$$\vspace*{-.2in}}

\noindent which compares
$x$'s
probability of occurrence conditioned on the appearance of an NPI with 
its
probability of occurrence overall.\footnote{\dld used an additional {\em distilled} score, but we found that the distilled score performed worse on Romanian.  \shrink{Our Romanian results for both \dld and our algorithm are
based on the undistilled scores
so as to make the comparison fair to \dld.}}

The method just outlined requires access to a list of NPIs.  \dld's
system used
a subset of
 John Lawler's carefully curated and ``moderately
complete'' 
list of English 
NPIs.%
\footnote{\url{http://www-personal.umich.edu/~jlawler/aue/npi.html}} The
resultant rankings of candidate
English
 \pos were judged to be of high quality.

\paragraph{The challenge in porting to other languages: cluelessness}

Can
the unsupervised approach of
 \dld be successfully applied to languages other than English? 
Unfortunately, for most
other languages, it does not seem that large, high-quality NPI lists
are available.

One might wonder whether one can circumvent the NPI-acquisition problem by simply translating 
a known English NPI list into the target language.  However,
NPI-hood need not be preserved under
translation \citep{ Richter+Rado+Sailer:08a}.
Thus, for most
languages, we
lack the critical clues that \dld depends on.

%% file: acl-short-results.tex
 \newcommand{\dmops}{DEops\xspace}
\newcommand{\pNPI}{pNPI\xspace}
\newcommand{\pNPIs}{pNPIs\xspace}
\newcommand{\sdld}{rDLD\xspace}

 \newcommand{\step}[1]{({\em #1})}
 \newcommand{\clues}{{\cal N}}%
\newcommand{\numpos}{n}
\newcommand{\numpnpis}{n_r}
 \newcommand{\ouroutput}{{\cal D}}

\section{Getting a clue}

In this section, we 
develop an iterative co-learning algorithm that
can extract \pos in the many languages where a high-quality NPI database is not available,
using Romanian as a case study.

\subsection{Data and evaluation paradigm}
We used Rada Mihalcea's corpus of $\approx$1.45 million sentences of raw Romanian newswire articles.

Note that  we cannot
evaluate impact on textual inference because, to our knowledge, no publicly available
textual-entailment
system 
or evaluation data for Romanian exists.  We therefore examine the system outputs
directly
 to determine whether the top-ranked items are actually \pos or not.  Our evaluation metric is precision at $k$ of a given system's ranked list of candidate \pos; 
it is not possible to evaluate recall since no list of Romanian \pos 
exists
 (a problem 
that
is precisely the motivation 
for
  this paper).

To evaluate the results, two native Romanian speakers labeled the system outputs as being ``DE'', ``not DE'' or ``Hard (to decide)''.
The labeling protocol, which was somewhat complex to prevent bias, is described \elsewhere
(\S7.1). 
The complete system output and
annotations are
publicly available at: {\dataurl}.

\subsection{Generating a seed set}\label{sec:seed}

Even though, as discussed
above, 
the translation of an NPI need not be an NPI, a preliminary review of the literature indicates that in many
languages, there is some NPI that can be translated as \ex{any} or
related forms like \ex{anybody}. 
Thus, with a small 
amount of
effort, one can
form a minimal NPI seed
set
for the \dld method
by using an appropriate target-language translation of \ex{any}.
For Romanian, we used
\ex{vreo} and \ex{vreun}, which are the feminine and masculine translations of English \ex{any}.

\input{acl-short-main-results-fig}

\subsection{\dld using the Romanian seed set}
We first check
whether \dld with the two-item seed set described 
in 
\S\ref{sec:seed} 
performs
 well on Romanian\shrink{; after all, if the results are good, then there
is no need 
for
new algorithms}.
In fact, the results are fairly poor: for example, the precision at 30 is below 50\%.  (See 
blue/dark bars in
figure
\numfigbarsnoiter\
\elsewhere for detailed results.)

This relatively unsatisfactory performance
may be a consequence 
of the 
very small
size of the NPI list employed,
 and 
 may
therefore indicate that it would be fruitful to investigate automatically
extending
our list of clues.

\subsection{Main idea: a co-learning approach}

Our main
 insight is that not only 
can NPIs
be used as clues for finding \pos, as shown by \dld, but
conversely, 
\pos (if known)
can potentially be used to discover new NPI-like clues,
which we refer to as {\em pseudo-NPIs} (or {\em \pNPIs} for short).
By ``NPI-like'' we mean, ``serve as 
possible indicators of the presence of \pos, regardless of whether they are actually restricted to 
negative contexts, 
as true NPIs are''.
For example, in English newswire, the words \ex{allegation} or \ex{rumor} tend to occur 
mainly 
in \dm contexts,
like  
\ex{\showop{denied}} or 
\ex{\showop{dismissed}},
 even though they are clearly not true NPIs (the sentence \ex{I heard a rumor} is fine).
Given this insight, we approach the problem using an {\em iterative co-learning} paradigm
that integrates the search for new \pos 
with a search for new  \pNPIs.  

First, we describe an algorithm that is the ``reverse''  of \dld (henceforth
{\em  \sdld}), in that it retrieves and ranks \pNPIs assuming a given list of \pos.  
Potential
\pNPIs are collected by extracting those words that appear in a \dm context (defined here,
to avoid the problems of parsing or scope determination,
 as the part of the sentence to the right of a \po, up to the first comma, semi-colon or end of sentence); these candidates  $x$ are then ranked by

{\small \vspace*{-.2in} $$\scorepnpis(x) :=  \frac
{\mbox{fraction of \dm contexts that contain $x$} }
{\mbox{relative frequency of $x$ in the corpus}}.
$$
}\vspace*{-.2in}

Then, our
{\em co-learning} algorithm
 consists of the iteration of the following two steps:

\begin{itemize}

\item \step{\dm learning} Apply \dld 
using a set $\clues$ of 
pseudo-NPIs
to retrieve a
list of candidate \pos ranked by $\scorepos$ (defined in Section \ref{sec:dld}).
Let $\ouroutput$ be the top $\numpos$  candidates in this list. 

\item  \step{\pNPI learning} Apply \sdld
 using the set $\ouroutput$ to retrieve a list of
  \pNPIs 
  ranked by $\scorepnpis$; extend $\clues$ with the top $\numpnpis$ \pNPIs in this list. Increment $\numpos$.
\end{itemize}
Here,  $\clues$ is initialized with
the NPI seed set.
At each iteration, we consider the output of the algorithm
to be the ranked list of \pos retrieved in 
the \dm-learning step. 
In our experiments, we initialized $\numpos$ to 10 and set $\numpnpis$ to 1.

\section{Romanian results}\label{sec:results}

Our results
  show
that there is indeed 
favorable synergy 
between
\dm-\oplong and \pNPI retrieval.  
Figure \ref{fig:nrops}  plots the number of correctly retrieved \pos in the top $k$ outputs at each 
iteration.
The point  at iteration $0$ corresponds to a datapoint already discussed above, namely, \dld applied to the two \ex{any}-translation NPIs.
Clearly, we see general substantial improvement over \dld, although the increases level off in later iterations. 
(Determining how to choose the optimal number of iterations is a subject for future research.)

Additional experiments, described {\elsewhere}
(\S7.\numsecmarginal),
suggest 
that \pNPIs 
can
even be more effective clues 
than a %
 noisy list of NPIs.  
 (Thus, a larger  seed set does not necessarily mean better performance.)
\pNPIs also have the advantage of being
derivable automatically, and might be worth investigating from a linguistic perspective in their own right.

%% file: acl-short-main-results-fig.tex
\begin{figure*}[ht]
\begin{tabular}{cc}
\includegraphics[width=3.25in,viewport= 100 235 570 557,clip]
{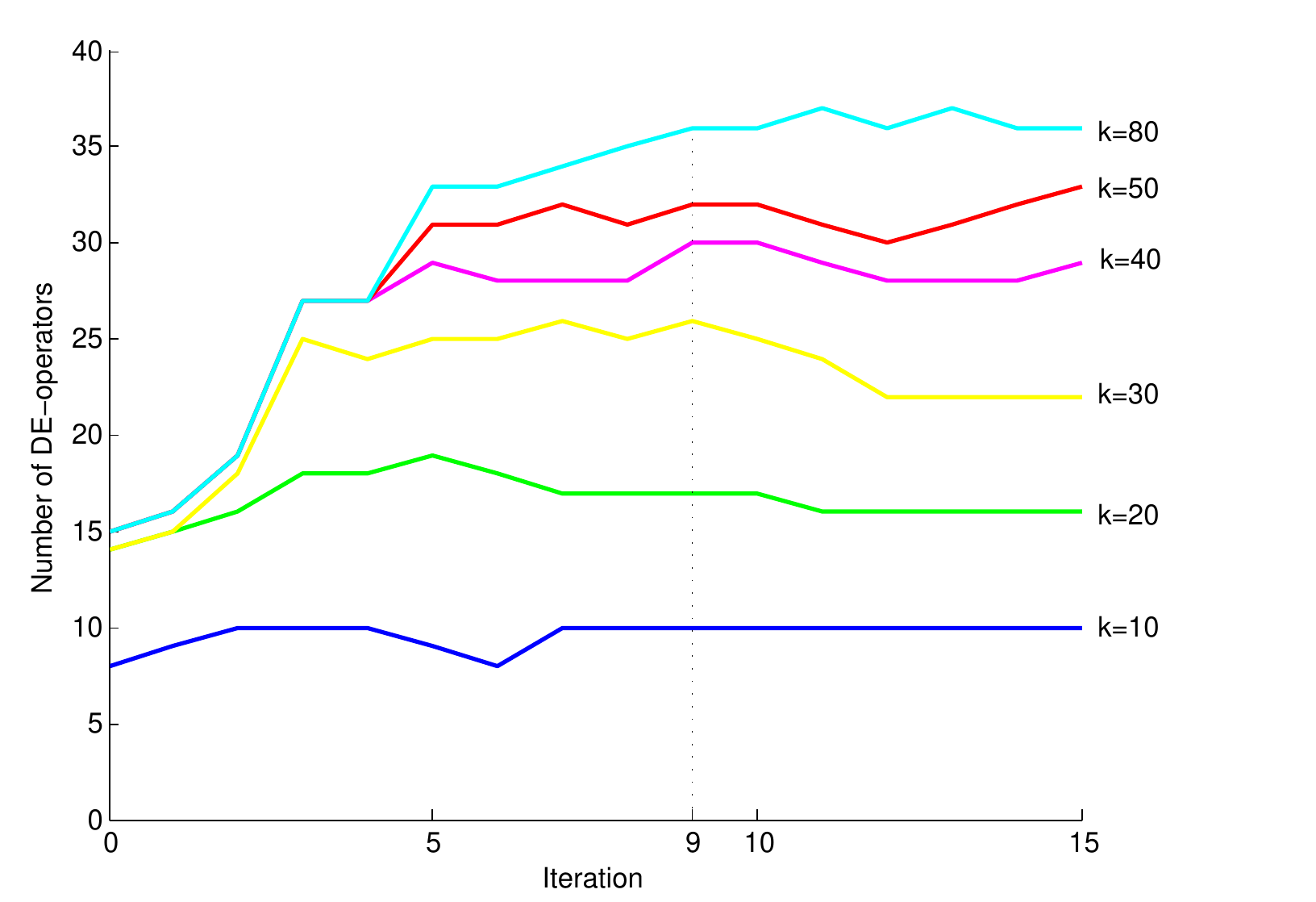}&
\includegraphics[width=3.25in,viewport= 100 235 570 557,clip]{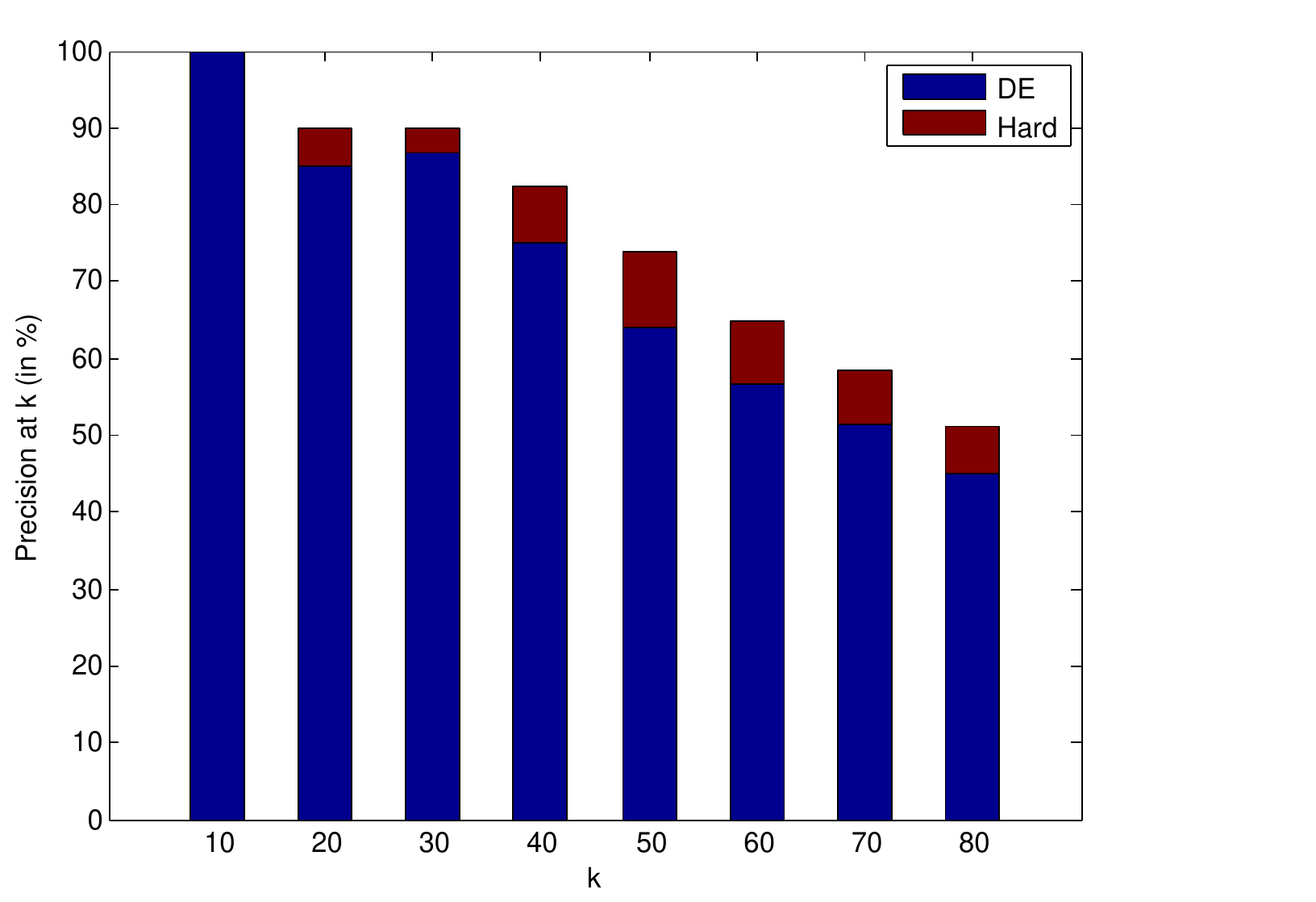} \\
\end{tabular}
\caption{
\small
\label{fig:nrops} 
{\bf Left:}  Number of \pos in the top $k$ results returned by the co-learning method at each iteration. Items labeled ``Hard'' are not included. 
Iteration 0 corresponds to \dld applied to \{\ex{vreo}, \ex{vreun}\}.
Curves for $k=60$ and $70$ omitted for clarity.
{\bf Right:} Precisions at $k$
for the results of the $9$th iteration.  The bar divisions are: DE 
(blue/darkest/largest) and Hard (red/lighter, sometimes non-existent).}
\end{figure*}

%% file: acl-short-multilingual.tex
\section{Cross-linguistic analysis}\label{sec:multilingual}

\input{acl-short-english}

\paragraph{Using translation}
Another interesting question is whether directly translating \pos from English is an alternative to our method.  First, 
we
emphasize that there exists no complete
list of English \pos
 (the largest available 
collection
is the one extracted by 
\dld
). Second,
we do not know whether
\pos in one language translate into \pos in another language.
Even if that 
were the case, and we
somehow had access to
ideal translations
of 
\dld's list,
there would still be considerable value in using our method: 
14 (39\%) of our top
36 highest-ranked Romanian \pos
for iteration 9
do not, according to 
the Romanian-speaking author,  
have English equivalents appearing on \dld's 
90-item list.
Some examples are: 
\ex{ab\c{t}inut} (abstained), \ex{criticat} (criticized)  and \ex{reac\c{t}ionat} (reacted).
\noindent Therefore, a significant fraction of the \pos derived by our co-learning algorithm would have been missed by the translation alternative even
under
 ideal conditions.

%% file: acl-short-english.tex
\vspace*{-.1in}

\paragraph{Applying our algorithm to English: connections to linguistic typology}
So far, we 
have
made no assumptions 
about the language on which our algorithm is applied.  A valid question is, 
does the quality of the results vary with choice of application language?
In particular, what happens if we run our algorithm on English?  

Note that in some sense, this is a perverse question:  
the motivation behind \shrink{the development of} our algorithm is the  non-existence of a high-quality  list of NPIs for the language in question, and English is essentially the only case that does not fit this description.  On the other hand, the fact that  \dld applied their method for extraction of \pos to English necessitates some form of comparison, for the sake of experimental completeness.

We thus ran our algorithm on the English BLLIP newswire corpus
{with seed set \{\ex{any}\}}.
We observe
that,
surprisingly, 
 the
iterative
 addition of \pNPIs   
has very little effect: the precisions at $k$
 are good at the beginning and stay about the same across iterations 
 (for details see figure
 \numfigenglish\
 in \elsewhere).  Thus, on English,  co-learning does not hurt performance, which is good news; but
unlike in Romanian, 
 it does not lead to improvements.

Why is English \ex{any} seemingly so ``powerful'', in contrast to Romanian, where iterating beyond the initial  \ex{any} translations leads to better results?
Interestingly,
findings from
 linguistic typology 
may shed some light on this issue.
\citet{Haspelmath:01a} compares the 
functions of
indefinite pronouns in 40 languages.
He shows that English is one of the 
minority
of languages  (11 out of 40){\footnote{
English\shrink{ (\ex{any})}, 
Ancash Quechua\shrink{ (\ex{-pis})},
Basque\shrink{ (\ex{i-})},
Catalan\shrink{ (\ex{cap})},
French\shrink{ (\ex{personne})}, %
Hindi/Urdu\shrink{ (\ex{bhii})},
Irish\shrink{ (\ex{dada})}, %
Portuguese\shrink{ (\ex{qualquer})}, 
Swahili\shrink{ (\ex{-ote})}, %
Swedish\shrink{ (\ex{n\aa gon})},
Turkish\shrink{ (\ex{herhangi})}.} }
in which there exists an
indefinite pronoun series that 
occurs in all (Haspelmath's) classes of
\dm contexts, 
 and thus can constitute
a sufficient seed
on its own.  
In the other languages  (including Romanian),\footnote{
{Examples:}
Chinese, \shrink{Dutch,} German, Italian, \shrink{Japanese,} Polish, 
 Serbian.} 
no indirect pronoun can serve as a sufficient seed.
So,
we expect
our method to be viable for all languages;
while
 the iterative discovery of \pNPIs is not necessary (although neither is it harmful)
 for the subset of languages for which a sufficient seed exists, such as 
 English,
it is essential for the languages for which, like Romanian, \ex{any}-equivalents do not suffice.

%% file: acl-short-conclusions.tex
\section{Conclusions} 
\vspace{-.1in}

We have introduced
the first method for discovering 
\dmlonghyph \opslong that is universally applicable.
Previous work on automatically detecting \pos assumed the existence of a high-quality collection of NPIs, which renders it inapplicable in most languages, where such a resource does not exist.  
We
overcome this limitation by employing a novel {\em co-learning} approach, and 
demonstrate
 its effectiveness on Romanian.

{Also,} we introduce the concept of {\em pseudo-NPIs}.
Auxiliary experiments described \elsewhere show that 
\mbox{\pNPIs} are actually more effective 
seeds than a noisy ``true'' NPI list.

Finally, we noted some cross-linguistic differences
in performance, 
and found an interesting connection between these 
differences
and Haspelmath's 
\citep{Haspelmath:01a} 
characterization of cross-linguistic variation in
the
 occurrence of indefinite pronouns.

%% file: esub-ack.tex
\paragraph{Acknowledgments}  
We thank Tudor Marian 
for serving as an annotator, 
Rada Mihalcea for access to the Romanian newswire corpus,
and
Claire Cardie,
Yejin Choi, 
Effi Georgala, 
Mark Liberman,
Myle Ott,  
 Jo\~ao Paula Muchado,
Stephen Purpura,
Mark Yatskar,
Ainur Yessenalina,
and the anonymous reviewers for their 
helpful
comments. Supported by NSF grant IIS-0910664.